\documentclass[letterpaper, 10 pt, conference]{ieeeconf}

\usepackage{amsmath}
\usepackage[noend]{algpseudocode}
\usepackage{verbatim}
\usepackage{makecell}
\usepackage{graphicx}
\usepackage{multirow}
\usepackage[ruled]{algorithm2e}

\IEEEoverridecommandlockouts 
\title{\LARGE \bf
Boosting Masked Face Recognition with Multi-Task ArcFace
}

\author{David Montero, Marcos Nieto, Peter Leskovsky and Naiara Aginako% <-this % stops a space
\thanks{David Montero, Marcos Nieto and Peter Leskovsky are with Vicomtech Foundation, Basque Research and Technology Alliance (BRTA), Mikeletegi 57, 20009 Donostia-San Sebasti\'an (Spain). Email: dmontero@vicomtech.org, mnieto@vicomtech.org, pleskovsky@vicomtech.org}
\thanks{Naiara Aginako is with the University of the Basque Country. Email: naiara.aginako@ehu.eus}
\thanks{The paper is under consideration at Pattern Recognition Letters}%
}

\begin{document}

\maketitle
\thispagestyle{empty}
\pagestyle{empty}

%%%%%%%%%%%%%%%%%%%%%%%%%%%%%%%%%%%%%%%%%%%%%%%%%%%%%%%%%%%%%%%%%%%%%%%%%%%%%%%%
\begin{abstract}
In this paper, we address the problem of face recognition with masks. Given the global health crisis caused by COVID-19, mouth and nose-covering masks have become an essential everyday-clothing-accessory. This sanitary measure has put the state-of-the-art face recognition models on the ropes since they have not been designed to work with masked faces. In addition, the need has arisen for applications capable of detecting whether the subjects are wearing masks to control the spread of the virus.
To overcome these problems a full training pipeline is presented based on the ArcFace work, with several modifications for the backbone and the loss function. From the original face-recognition dataset, a masked version is generated using data augmentation, and both datasets are combined during the training process. The selected network, based on ResNet-50, is modified to also output the probability of mask usage without adding any computational cost. Furthermore, the ArcFace loss is combined with the mask-usage classification loss, resulting in a new function named Multi-Task ArcFace (MTArcFace).
Experimental results show that the proposed approach highly boosts the original model accuracy when dealing with masked faces, while preserving almost the same accuracy on the original non-masked datasets. Furthermore, it achieves an average accuracy of 99.78\% in mask-usage classification.

\end{abstract}
%%%%%%%%%%%%%%%%%%%%%%%%%%%%%%%%%%%%%%%%%%%%%%%%%%%%%%%%%%%%%%%%%%%%%%%%%%%%%%%%

\section{Introduction}

% 1. face recognition muy importante porque blabla
In recent years, advances in the field of face recognition have made it one of the most reliable biometric techniques among other existing techniques such as fingerprint recognition, hand geometry or iris scanning \cite{biotechs1, biotechs2}. Furthermore, compared to its alternatives, facial recognition has the following advantages:
\begin{itemize}
\item It is a more affordable solution than other alternatives such as fingerprint or iris scanners, as it only needs a mono camera as a sensor. In addition, several cameras may be connected to a single processing unit to even more reduce hardware costs.
\item The verification can be done remotely; there is no need for the user to interact with the sensor.
\item The sensor can be hidden, which can be very useful for security or aesthetic reasons.
\end{itemize}

All these features make face recognition the best choice for most of the applications based on human re-identification.

% 2. COVID causa mascarillas causa face recognition funciona mal
However, face recognition also has weaknesses. Current state-of-the-art methods are based on deep learning models that extract biometric feature vectors from the detected face images. These detected faces may have different orientations, lighting conditions, partial occlusions, low resolution, noise, etc., which can affect the robustness of the feature vectors \cite{Tan2010}.

Most of these negative conditions can often be eliminated by selecting the correct hardware location and requirements and by preprocessing face images \cite{Chaudhari2010}, but others such as partial occlusions caused by clothing accessories cannot be avoided. This particular issue have recently become a major challenge in the field of face recognition, especially since the global health crisis originated by COVID-19 has caused medical face masks to become an everyday-clothing-accessory.

%%% mete fotito de cara con y sin mascarilla con puntitos enseñando la pérdida de datos biométricos

The use of a mouth and nose-covering mask makes the face recognition models to lose about half of the useful biometric information. Since they have been designed to work with the whole face information, the quality of the feature vectors extracted from masked faces is compromised and the accuracy of the re-identification process decreases considerably, as stated in \cite{maskedEffect}. In fact, NIST agency recently presented a study \cite{maskedEffect1} where they examined 89 major commercial facial recognition algorithms. The results showed error rates of between 5\% and 50\% in matching photos of the same person with and without a mask.

% mnieto: no entiendo bien la última frase. ¿Sin mask el error medio es 5%, y con mask el error medio es 50%? Si es así, ponlo así please.

% 3. que se está haciendo para solucionar esto
While masked face detection has been widely studied and several robust solutions have been presented \cite{retinamask,maskedDet,maskedDet1}, masked face recognition remains an under-researched topic. In the last months, several masked face datasets \cite{Cabani2020MaskedFaceNetA,wang2020masked} and tools \cite{Anwar2020MaskedFR,maskedgan2} for generating synthetic data have been released. In addition, some methods trying to tackle this issue using different approaches have been presented \cite{maskedDCR,maskedgan2,maskedCrop}. Nevertheless, there is still much research to be done about this topic.

% mnieto: Quizá un pequeño paréntesis al principio para distinguir lo que es "face detection" vs "face recognition" podría ayudar a un lector no experto. E.g. "While masked face detection (the ability to determine the region of the image that contains a face with a mask) [...] masked face recognition (the ability to determine the identity of the person wearing the mask compared to a reference or database)"

% 4. proponer solución y lo hemos probado asi
To contribute to this task, we propose an approach based on the ArcFace work presented by Deng et al. \cite{arcface} with several modifications for the backbone and the loss function. From the original face-recognition dataset, we generate a masked version using data augmentation, and we combine both datasets during the training process. We modify the selected network, based on ResNet-50 \cite{resnet,He2016}, to also output the probability that a face is wearing a mask without adding any additional computational cost. Furthermore, we combine the ArcFace loss with the mask-usage classification loss, resulting in a new function named Multi-Task ArcFace (MTArcFace).

Experimental results with non-masked and masked face-recognition validation datasets show that the proposed approach highly boosts the model accuracy when dealing with masked face recognition, while preserving almost the same accuracy on the non-masked datasets. Furthermore, the model achieves an accuracy of 99.78\% in mask-usage classification.

% 6. índice
The rest of the paper is organized as follows. First, we present a review of the related work in Section \ref{sec:related}. Section \ref{sec:method} describes the proposed method. We provide the experimental results in Section \ref{sec:experiments}. Finally, conclusions are given in section \ref{sec:conclusions}.

\section{Related Work} \label{sec:related}
\subsection{Face Recognition} \label{related-facerec}
State-of-the-art face recognition algorithms are based on deep learning models. These models learn to extract the important features from a face image an embed them into an n-dimensional vector with small intra-class and large inter-class distance.

These models are trained mainly following two approaches. The first one consists on training a multi-class classifier considering one class for each identity in the training dataset, normally using a softmax function \cite{arcface, sphereface}. In the second one, the embedding is learnt directly, comparing the results of different inputs to minimize the intra-class distance and to maximize the inter-class distance, for example using the triplet loss \cite{facenet}.

Both softmax-loss-based and triplet-loss-based models suffer from face-mask occlusions in terms of accuracy, as reported by \cite{maskedEffect} and \cite{maskedEffect1}. However, as stated in \cite{arcface}, triplet-loss-based models require a data preparation step prior to the training phase, in order to select the triplets correctly. For this reason, we decided to address the problem using a softmax-loss approach. More specifically, we selected ArcFace \cite{arcface} as our baseline, since it has been proven to be the approach that reports the best results for the face recognition task.

\subsection{Masked Face Recognition}

Since the rise of COVID-19, several works have been presented in order to solve masked face recognition task. The proposed methods tackle the problem following different approaches that can be categorized in three groups. The first group uses generative adversarial networks (GAN) to unmask faces prior to feeding them to the face recognition model \cite{maskedgan1,maskedgan2}. Using this approach it is not necessary to retrain the recognition model. However, the reconstructed faces are synthetic and their reliability depends on the quality of the data, the network and the training process. In addition, the process of removing the mask noticeably increases the computation time.

The approach adopted by the second group consists of extracting features only from the upper part of the face \cite{maskedCrop}. As the processed region of the face is smaller, the trained network performs faster. Nevertheless, this causes an important drop of information when dealing with unmasked faces, so it is not suitable for applications mixing both use cases.

Finally, the last group tackles the problem training the face recognition network with a combination of masked and unmasked faces \cite{Anwar2020MaskedFR, maskedDCR}. In \cite{Anwar2020MaskedFR} they combine the VGG2 dataset \cite{vgg2} with augmented masked faces and train the model following the original pipeline described in FaceNet \cite{facenet}. This way, the model learns to distinguish when a face is wearing a mask and to trust more in the features of the upper half of the face, but still extracts information from the whole face. On the other hand, Geng et al. \cite{maskedDCR} define two centers for each identity which correspond to the full face images and the masked face images respectively. They use Domain Constrained Ranking for forcing the feature of masked faces getting closer to its corresponding full face center and vice-versa.

\begin{figure*}[t]
    \centering
    \includegraphics[width=\linewidth]{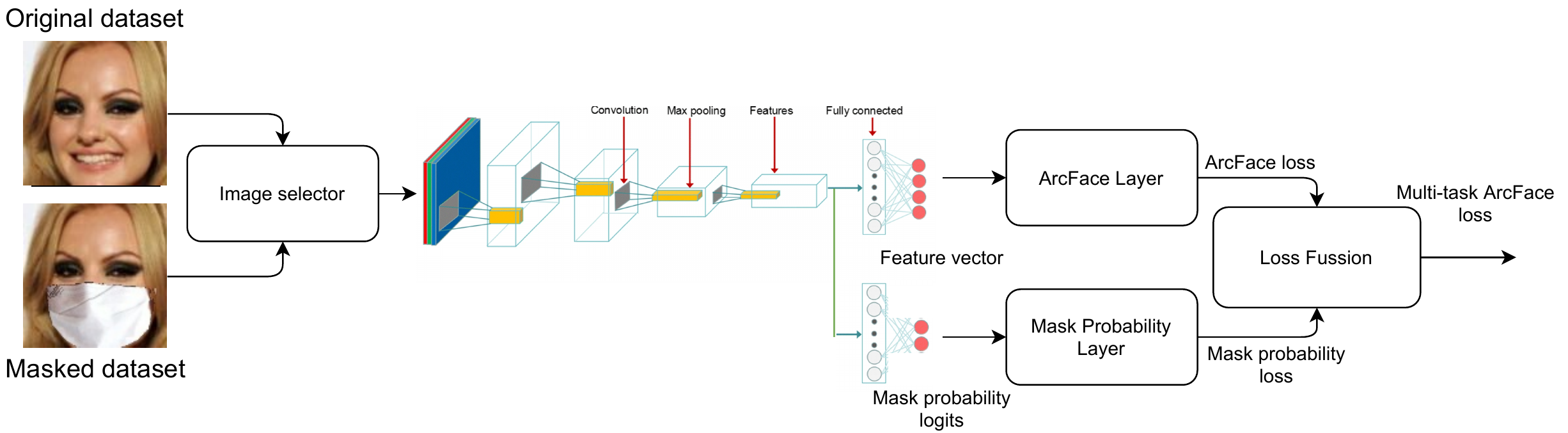}
    \caption{Illustration of the proposed training pipeline. The image selector decides whether the next input image should be masked or not. The trained network is modified to output also the probability that the face is wearing a mask.}
    \label{fig:pipeline}
\end{figure*}

The method proposed in this work belongs to the third group, but using ArcFace \cite{arcface} as the baseline model. First, we generate a masked version of a face recognition dataset using data augmentation. Then, during the training process, both datasets are shuffled separately using the same seed and, for every new face image selected for the input batch, we decide whether the image is taken from the original or the masked dataset with a probability of 50\%. Furthermore, we take advantage of knowing to which dataset the face belongs to and modify the original network to output the probability that a face is wearing a mask without additional computational cost.

% mnieto: "we take advantage of knowing to which dataset the face belongs to" ¿Se puede refrasear? Me suena raro.

\subsection{Masked Face Datasets} \label{related-datasets}
For the methods belonging to the first and the second group previously described, there is a need of masked face datasets. Some recent works have contributed to this task. For instance, Geng et al. \cite{maskedDCR} present a dataset where each identity has masked and full face images with various orientations. However, the dataset contains only 11,615 images and 1,004 identities, which is not enough data for training a complex network such as ResNet-50 \cite{resnet,He2016}. In \cite{Cabani2020MaskedFaceNetA}, the authors present a dataset composed of 137,016 masked faces divided in two groups: correctly and incorrectly masked. Nevertheless, the dataset does not contain information about the identity of any of the subjects, so it cannot be used for the face recognition task. In \cite{wang2020masked}, two additional datasets are presented: Real-world Masked Face Recognition Dataset (RMFRD), with 95,000 images and 525 identites, and Simulated Masked Face Recognition Dataset (SMFRD), with 500,000 and 10,000 subjects. Although the latter dataset contains a great number of samples, it is not yet sufficient to train a complex network, for example if we compare it with MS1MV2 dataset used in ArcFace \cite{arcface}, which contains 5.8 million images and 85,000 identities.

On the other hand, Anwar and Raychowdhury \cite{Anwar2020MaskedFR} present a tool for masking faces in images. It uses a face landmarks detector to identify the face tilt and six key features of the face necessary for adjusting and applying a mask template. This tool supports different types and colors of masks. In this work, we use this tool to generate a masked version of the face recognition datasets used for training and evaluation.

\section{Proposed Method} \label{sec:method}
% 1. introduccion del problema
\subsection{Problem Definition}
We consider the problem of facial recognition of subjects who may or may not wear masks. As we do not know if the subject is wearing a mask, the network must perform well in both cases.
%%% foto del problema?
% 2. enfoque
To solve this problem, we aim at increasing the accuracy of the face recognition network when dealing with masked faces, while preserving as much as possible the original accuracy with non-masked faces. In order to achieve this, the network must learn if the subject is wearing a mask to decide which facial features can be trusted in each case. We take advantage of this fact and modify the network so it also outputs the probability that the subject is wearing a mask.

% 3. pipeline entrenamiento
\subsection{Training Pipeline}
We decide to generate a masked twin dataset from the original one and to combine them during the training process. Both datasets are shuffled separately using the same seed and, for every new face image selected for the input batch, we decide whether the image is taken from the original or the masked dataset with a probability of 50\%. As mentioned in Section \ref{related-facerec}, we use ArcFace \cite{arcface} as the baseline work for two reasons: it uses a softmax-loss-based methodology, which does not require an exhaustive training-data-preparation stage; and it has been proven to be the approach that reports the best results for the original face recognition task. Thus, we select the dataset recommended in their work MS1MV2 as the training dataset, which is a refinement of MS-Celeb-1M \cite{msceleb}, which contains 5.8M images and 85,000 identities. An illustration of the proposed training pipeline is shown in Figure \ref{fig:pipeline}.

\begin{figure*}[t]
    \centering
    \includegraphics[width=\linewidth]{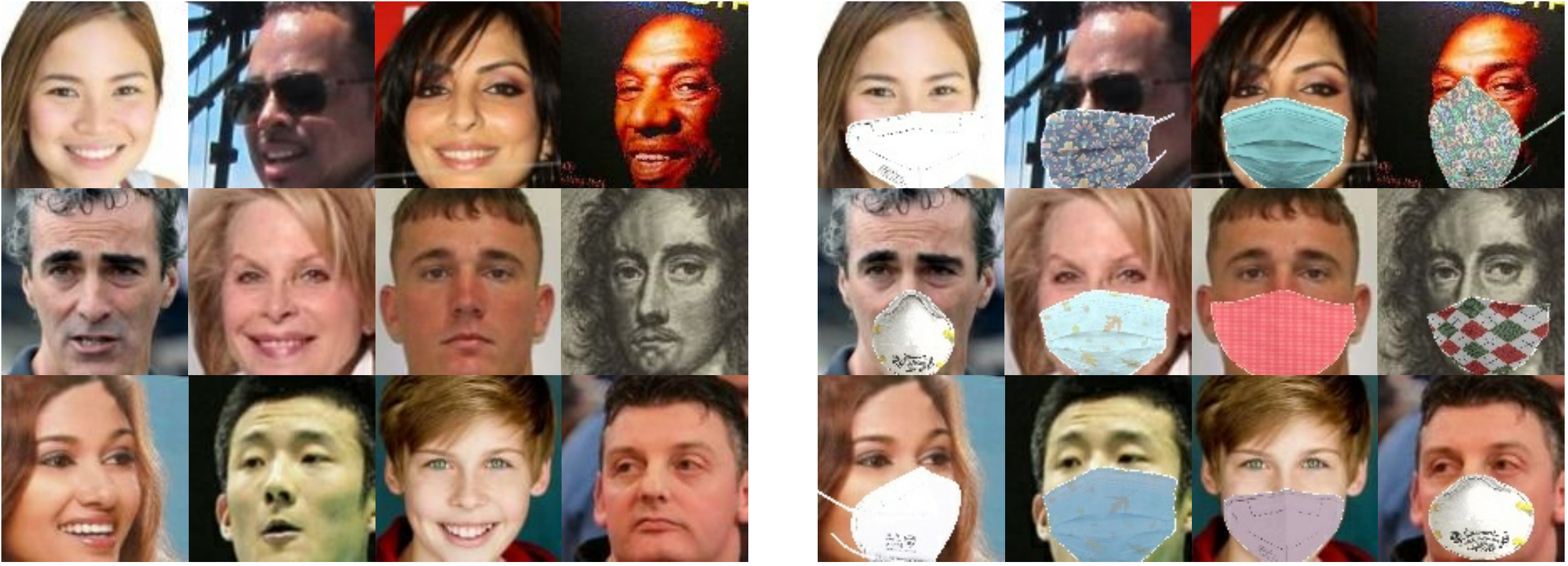}
    \caption{Some examples of the training faces and their corresponding masked version generated with the MaskTheFace tool \cite{Anwar2020MaskedFR}}.
    \label{fig:training_data}
\end{figure*}

% 4. generación de dataset
For the generation of the masked version of the dataset, as discussed in section \ref{related-datasets}, we use the tool MaskTheFace \cite{Anwar2020MaskedFR}. The types of masks considered are surgical, surgical green, surgical blue, N95, cloth and KN95. The type mask is selected randomly and there is a probability of 50\% of applying a random color and a probability of 50\% of applying a random texture. Some examples of the generated faces are shown in Figure \ref{fig:training_data}.

% We believe that the type of the mask it is not important for the network, as it only needs to learn that when the lower region of the face is covered by a mask, those features are not reliable. Thus, we generate the dataset using only the surgical masks with random colors, as shown in Figure \ref{fig:training_data}.

% 5. eleccion de arquitectura y modificacion
We select LResNet-50 as the backbone among all the network architectures tested in the ArcFace repository as it is the one with the best trade-off between the accuracy and the number of parameters. More specifically, we use our own implementation of the network in TensorFlow deep learning framework, publicly available in a GitHub repository \cite{dmonteroTF2}.

% mnieto: ¿Algún motivo? Por ejemplo "we use our own implementation... because it..."

Starting from this network, we add another dense layer parallel to the one used to generate the feature vector, just after the dropout layer, as shown in Figure \ref{fig:pipeline}. The new dense layer generates an output with two floats, which correspond to the scores related to the probability that the face is masked or not, respectively. This way, we force the network to learn when a face is wearing a mask, information that will also be used by the layer that generates the feature vector.

% 6. modificación de loss function
Thus, from the modified network we obtain two outputs, the logits (unnormalized predictions) of the ArcFace layer ($logits_{ArcFace}$) and the logits of the new dense layer ($logits_{Mask}$). To extract the combined error from both logits, we start by generating the ArcFace loss ($loss_{ArcFace}$) in the same way as in \cite{dmonteroTF2}:

\begin{equation}
loss_{ArcFace} = crossEnt(Softmax(logits_{ArcFace}, labels_{ID})
\label{eq:arcface_loss}
\end{equation}

\noindent Next, we calculate the loss associated with the probability of wearing a mask ($loss_{Mask}$) by applying the softmax activation function on the logits and cross-entropy with the labels:

\begin{equation}
loss_{Mask} = crossEnt(Softmax(logits_{Mask}), labels_{ID})
\label{eq:mask_loss}
\end{equation}

\noindent The Multi-Task ArcFace loss ($loss_{MTArcFace}$) is obtained by adding these two losses. However, to reduce the impact of $loss_{Mask}$ and give more importance to the ArcFace loss, we use the logarithm of $loss_{Mask}$ instead of the original value:

\begin{equation}
loss_{MTArcFace} = loss_{ArcFace} + log(loss_{Mask} + 1.0)
\label{eq:mtarcface_loss}
\end{equation}

\noindent Finally, we add the regularization loss (as in the original implementation) to compute the total loss that will be used for the optimization:

\begin{equation}
loss_{total} = loss_{MTArcFace} + loss_{regularization}
\label{eq:total_loss}
\end{equation}

\begin{figure}[t]
    \centering
    \includegraphics[width=\linewidth]{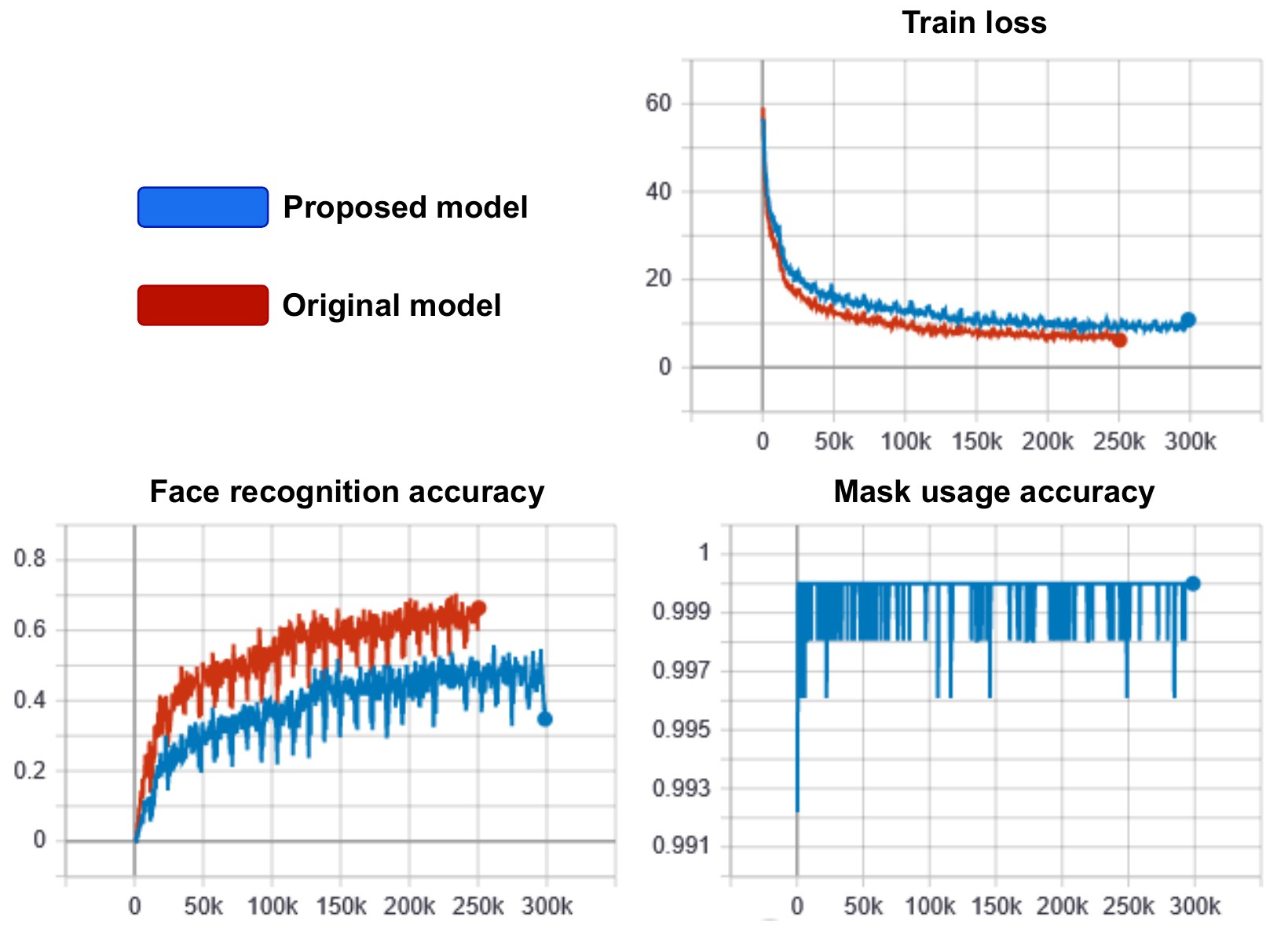}
    \caption{Training curves for the proposed model and the original ArcFace model. The X axis represents the training steps}.
    \label{fig:training_curves}
\end{figure}

% 7. parámetros de entrenamiento
We train the model using 2 Tesla V100 GPUs with a total batch size of 512 and for 300k steps. We use the SGD optimizer with a momentum of 0.9 and an initial learning rate of 0.0015. The learning rate is reduced by a factor of 0.3 in steps 120k, 200k and 280k. The rest of the parameters of the network remain the same as in the original implementation. In Figure \ref{fig:training_curves}, we show the training loss curve and the face-recognition and mask-usage accuracy curves, compared to those of the original model.

% mnieto: Usamos 300k, y en otros números grandes he visto 10,000. También algún 5.8M. La k y la M son informales.

\section{Experiments} \label{sec:experiments}
In this section, we present the results of a series of experiments aimed at demonstrating the capabilities of the proposed method. We divide the experiments into two groups: identity verification and mask-usage verification.

\subsection{Identity Verification}
The first group of experiments are conducted to measure the performance improvement of the proposed method in the face verification task when dealing with masked faces. For measuring this increase, we use the original model as the baseline to compare the results.

\begin{figure*}[t]
    \centering
    \includegraphics[width=\linewidth]{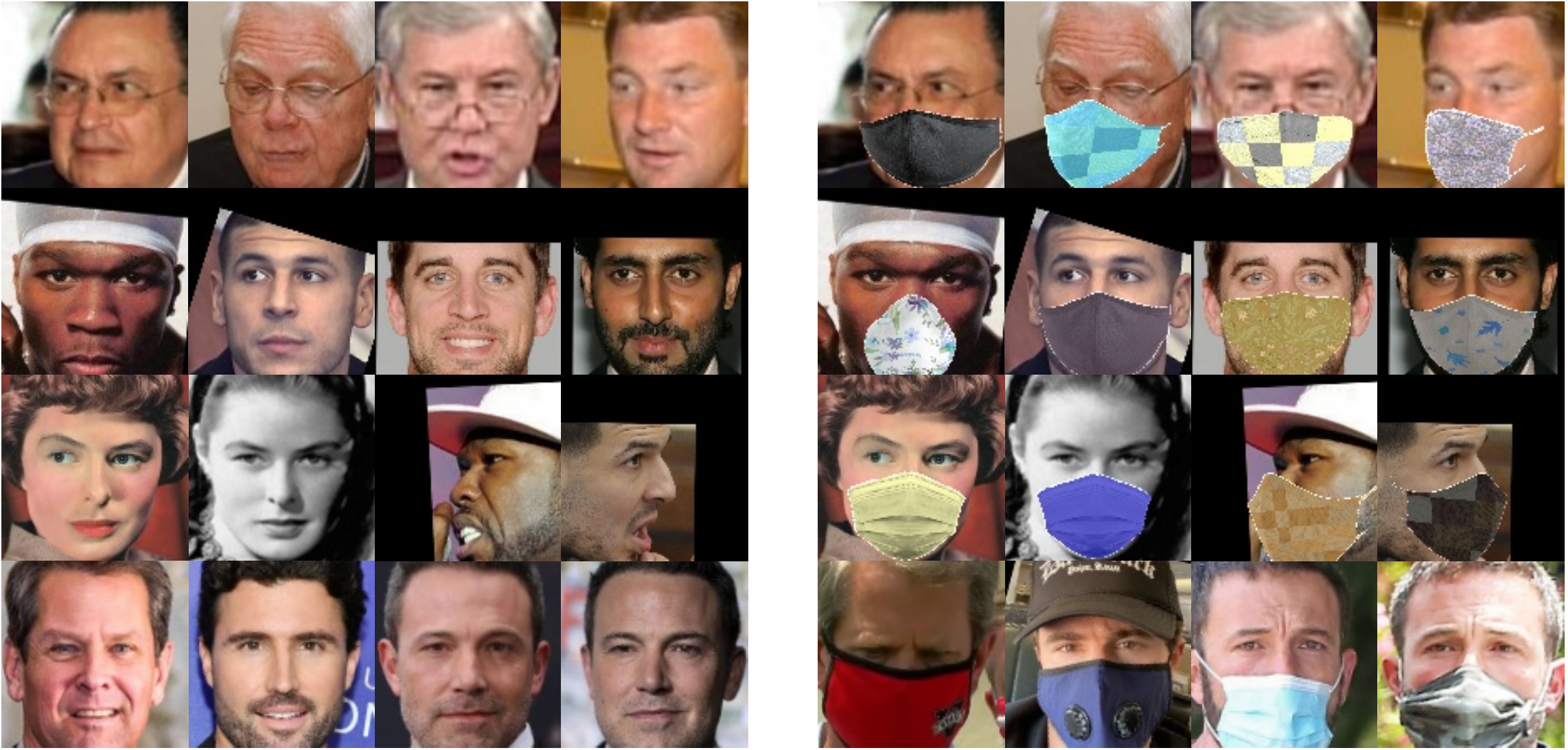}
    \caption{Some examples of the faces of the evaluation datasets and their masked versions. The first row belongs to LFW \cite{lfw}, the second row to CFP \cite{cfp}, the third row to AGEDB \cite{agedb} and the last row to MFR2 \cite{Anwar2020MaskedFR}}.
    \label{fig:evaluation_data}
\end{figure*}

% 1. datasets con mascarilla
For the verification task, we generate masked versions of 3 well-known face recognition datasets, also used in \cite{arcface} for evaluating the original models:
\begin{itemize}
    \item Labeled Faces in the Wild (LFW) \cite{lfw}: public benchmark for face verification containing 13,233 images from 5,749 people. The associated face verification task includes 6,000 comparisons.
    \item Celebrities in Frontal-Profile in the Wild (CFP) \cite{cfp}: contains faces from 500 celebrities in frontal and profile views. Two verification protocols are presented for this dataset: one comparing only frontal faces (CFP\_FF), and the other comparing frontal and profile faces (CFP\_FP). Each of the protocols are composed of 7,000 comparisons. We consider both protocols for our experiments.
    \item Agedb \cite{agedb}: the first manually collected, in-the-wild age database. Contains 16,488 images from 568 celebrities at different ages. Contains four verification protocols where the compared faces have an age difference of 5, 10, 20 and 30 years respectively. We select the last protocol for the experiments (AgeDB\_30, as it is the most challenging one. It contains 6,000 comparisons.
\end{itemize}

In addition, we also consider for the experiment the masked face dataset MFR2 described in \cite{Anwar2020MaskedFR}, with 269 real-world face images from 53 celebrities, where the 64\% of the faces wear a mask. The associated verification process is composed of 848 comparisons. Some examples of the images of the different datasets considered for the experiment are shown in Figure \ref{fig:evaluation_data}.

The results of the experiment are presented in Table \ref{tab:masked_verification}. It can be observed that the proposed method largely outperforms the original model in the face verification task when dealing with masked faces. This increase in performance is more evident with profile images, where the amount of information of the face available is reduced, as is the case with CFP\_FP, where the proposed model is almost a 12\% more accurate than the original.

\begin{table}[t]
\caption{Comparison of the verification performance (\%) with the masked datasets between the proposed method and the original ArcFace model.}
\label{tab:masked_verification}
\begin{center}
 \begin{tabular}{|c||c|c|}
 \hline
 Dataset & Proposed Method & Original model \\
 \hline\hline
 Masked LFW & 98.92 & 94.75 \\ 
 \hline
 Masked CFP\_FF & 98.33 & 92.73 \\ 
 \hline
 Masked CFP\_FP & 88.43 & 76.81 \\ 
 \hline
 Masked AGEDB\_30 & 93.17 & 90.53 \\ 
 \hline
 MFR2 & 99.41 & 97.17 \\ 
 \hline
\end{tabular}
\end{center}
\end{table}

% 2. datasets originales
We also want to test the accuracy of the new model when recognizing non-masked faces, to check whether it has been a significant drop of performance. Thus, we repeat the previous experiment with the original non-masked datasets and compare the results with those achieved by the original model. The results, exposed in Table \ref{tab:original_verification}, show that there is indeed a drop of performance for the new model, but that it is not significant (less than a 2\% in the worst case). Furthermore, this drop in performance is much less than the gain obtained with masked faces. For example, in the case of CFP\_FP, the model accuracy with masked faces increases almost a 12\%, while its accuracy with non-masked faces decreases less than a 2\%.

\begin{table}[t]
\caption{Comparison of the verification performance (\%) with the original datasets between the proposed method and the original ArcFace model.}
\label{tab:original_verification}
\begin{center}
 \begin{tabular}{|c||c|c|}
 \hline
 Dataset & Proposed Method & Original model \\
 \hline\hline
 LFW & 99.45 & 99.62 \\ 
 \hline
 CFP\_FF & 99.40 & 99.70 \\ 
 \hline
 CFP\_FP & 92.27 & 93.81 \\ 
 \hline
 AGEDB\_30 & 95.02 & 96.90 \\
 \hline
\end{tabular}
\end{center}
\end{table}

\subsection{Mask-Usage Verification}
% 3. probabilidad de mascarilla
Finally, we want to analyze the performance of the mask-usage probability output added to the proposed method. For this task, we run the model with all the faces contained in every masked and non-masked dataset used in the previous experiments. For each face we check whether the mask-usage probability estimated by the model is correct or not with a threshold of 0.5. Table \ref{tab:masked_verification} shows the results of the experiment. For each dataset, the model achieves nearly 100\% accuracy. Again, the worst result is achieved for the CFP\_FP dataset (98.82\%) due to the profile faces. We believe that this is due to the fact that the training dataset does not contain enough profile faces. In any case, the model achieves an average accuracy of 99.78\% across all datasets, so its effectiveness for this task is demonstrated.
\begin{table}[t]
\caption{Mask-Usage verification performance (\%) of the proposed method.}
\label{tab:mask_verification}
\begin{center}
 \begin{tabular}{|c||c|}
 \hline
 Dataset & Accuracy \\
 \hline\hline
 LFW & 99.99 \\ 
 \hline
 CFP\_FF & 99.99 \\ 
 \hline
 CFP\_FP & 98.82 \\ 
 \hline
 AGEDB\_30 & 99.97 \\
 \hline
 Masked LFW & 99.98 \\ 
 \hline
 Masked CFP\_FF & 99.98 \\ 
 \hline
 Masked CFP\_FP & 99.70 \\ 
 \hline
 Masked AGEDB\_30 & 99.99 \\ 
 \hline
\end{tabular}
\end{center}
\end{table}

\section{Conclusions} \label{sec:conclusions}
In this work, we have presented a full-training pipeline for ArcFace-based face-recognition models to adapt them for working with masked faces. This pipeline includes the generation of a synthetic masked dataset from the original training dataset. Furthermore, we have taken advantage of knowing to which dataset the face belongs to and modified the original network to output the probability that a face is wearing a mask without additional computational cost. As a result, we have created a new loss function to teach the network to extract vectors of good quality and reliable mask-usage probabilities called Multi-Task ArcFace. Experimental results with multiple masked and non-masked datasets have demonstrated that the proposed method highly boosts the performance of the model when recognizing masked faces, while suffering just a small drop in performance with non-masked faces. Furthermore, it has also been demonstrated its effectiveness for the mask-usage verification task with an average performance of 99.78\% of accuracy across all datasets.

Future work will focus on extending the applicability of this method to other types of occlusions, such as eyes-masked faces. In addition, we will also study the possibility of adding a new output to the model to classify if the subject is wearing a mask correctly or if it is wearing it under its nose or its mouth.

\section{Acknowledgements}

This paper is supported by European Union’s Horizon 2020 research and innovation programme under grant agreement No 883341, project GRACE (Global Response Against Child Exploitation).

%%%%%%%%%%%%%%%%%%%%%%%%%%%%%%%%%%%%%%%%%%%%%%%%%%%%%%%%%%%%%%%%%%%%%%%%
\bibliographystyle{IEEEtran}
\bibliography{Biblio}

\begin{thebibliography}{10}
\providecommand{\url}[1]{#1}
\csname url@samestyle\endcsname
\providecommand{\newblock}{\relax}
\providecommand{\bibinfo}[2]{#2}
\providecommand{\BIBentrySTDinterwordspacing}{\spaceskip=0pt\relax}
\providecommand{\BIBentryALTinterwordstretchfactor}{4}
\providecommand{\BIBentryALTinterwordspacing}{\spaceskip=\fontdimen2\font plus
\BIBentryALTinterwordstretchfactor\fontdimen3\font minus
  \fontdimen4\font\relax}
\providecommand{\BIBforeignlanguage}[2]{{%
\expandafter\ifx\csname l@#1\endcsname\relax
\typeout{** WARNING: IEEEtran.bst: No hyphenation pattern has been}%
\typeout{** loaded for the language `#1'. Using the pattern for}%
\typeout{** the default language instead.}%
\else
\language=\csname l@#1\endcsname
\fi
#2}}
\providecommand{\BIBdecl}{\relax}
\BIBdecl

\bibitem{biotechs1}
D.~Mou, \emph{Fundamentals and Advances in Biometrics and Face
  Recognition}.\hskip 1em plus 0.5em minus 0.4em\relax Berlin, Heidelberg:
  Springer Berlin Heidelberg, 2010, ch.~2, pp. 13--70.

\bibitem{biotechs2}
N.~Anot and K.~Singh, ``A review on biometrics and face recognition
  techniques.'' \emph{International Journal of Advanced Research}, vol.~4, pp.
  783--786, 05 2016.

\bibitem{Tan2010}
X.~{Tan} and B.~{Triggs}, ``Enhanced local texture feature sets for face
  recognition under difficult lighting conditions,'' \emph{IEEE Transactions on
  Image Processing}, vol.~19, no.~6, pp. 1635--1650, June 2010.

\bibitem{Chaudhari2010}
S.~T. {Chaudhari} and A.~{Kale}, ``Face normalization: Enhancing face
  recognition,'' in \emph{2010 3rd International Conference on Emerging Trends
  in Engineering and Technology}, Nov 2010, pp. 520--525.

\bibitem{maskedEffect}
N.~{Damer}, J.~H. {Grebe}, C.~{Chen}, F.~{Boutros}, F.~{Kirchbuchner}, and
  A.~{Kuijper}, ``The effect of wearing a mask on face recognition performance:
  an exploratory study,'' in \emph{2020 International Conference of the
  Biometrics Special Interest Group (BIOSIG)}, 2020, pp. 1--6.

\bibitem{maskedEffect1}
``Nist finds flaws in facial checks on people with covid masks,''
  \emph{Biometric Technology Today}, vol. 2020, no.~8, p.~2, 2020.

\bibitem{retinamask}
M.~Jiang, X.~Fan, and H.~Yan, ``Retinamask: A face mask detector,'' 2020.

\bibitem{maskedDet}
M.~Loey, G.~Manogaran, M.~H.~N. Taha, and N.~E.~M. Khalifa, ``A hybrid deep
  transfer learning model with machine learning methods for face mask detection
  in the era of the covid-19 pandemic,'' \emph{Measurement}, vol. 167, p.
  108288, 2021.

\bibitem{maskedDet1}
S.~Lin, L.~Cai, X.~Lin, and R.~Ji, ``Masked face detection via a modified
  lenet,'' \emph{Neurocomputing}, vol. 218, pp. 197 -- 202, 2016.

\bibitem{Cabani2020MaskedFaceNetA}
A.~Cabani, K.~Hammoudi, H.~Benhabiles, and M.~Melkemi, ``Maskedface-net - a
  dataset of correctly/incorrectly masked face images in the context of
  covid-19,'' \emph{ArXiv}, vol. abs/2008.08016, 2020.

\bibitem{wang2020masked}
Z.~Wang, G.~Wang, B.~Huang, Z.~Xiong, Q.~Hong, H.~Wu, P.~Yi, K.~Jiang, N.~Wang,
  Y.~Pei, H.~Chen, Y.~Miao, Z.~Huang, and J.~Liang, ``Masked face recognition
  dataset and application,'' 2020.

\bibitem{Anwar2020MaskedFR}
A.~Anwar and A.~Raychowdhury, ``Masked face recognition for secure
  authentication,'' \emph{ArXiv}, vol. abs/2008.11104, 2020.

\bibitem{maskedgan2}
N.~{Ud Din}, K.~{Javed}, S.~{Bae}, and J.~{Yi}, ``A novel gan-based network for
  unmasking of masked face,'' \emph{IEEE Access}, vol.~8, pp. 44\,276--44\,287,
  2020.

\bibitem{maskedDCR}
M.~Geng, P.~Peng, Y.~Huang, and Y.~Tian, ``Masked face recognition with
  generative data augmentation and domain constrained ranking,'' in
  \emph{Proceedings of the 28th ACM International Conference on Multimedia},
  ser. MM '20.\hskip 1em plus 0.5em minus 0.4em\relax New York, NY, USA:
  Association for Computing Machinery, 2020, p. 2246–2254.

\bibitem{maskedCrop}
W.~Hariri, ``Efficient masked face recognition method during the covid-19
  pandemic,'' 2020.

\bibitem{arcface}
J.~Deng, J.~Guo, N.~Xue, and S.~Zafeiriou, ``Arcface: Additive angular margin
  loss for deep face recognition,'' in \emph{Proceedings of the IEEE/CVF
  Conference on Computer Vision and Pattern Recognition (CVPR)}, June 2019.

\bibitem{resnet}
D.~{Han}, J.~{Kim}, and J.~{Kim}, ``Deep pyramidal residual networks,'' in
  \emph{2017 IEEE Conference on Computer Vision and Pattern Recognition
  (CVPR)}, 2017, pp. 6307--6315.

\bibitem{He2016}
K.~{He}, X.~{Zhang}, S.~{Ren}, and J.~{Sun}, ``Deep residual learning for image
  recognition,'' in \emph{2016 IEEE Conference on Computer Vision and Pattern
  Recognition (CVPR)}, June 2016, pp. 770--778.

\bibitem{sphereface}
W.~{Liu}, Y.~{Wen}, Z.~{Yu}, M.~{Li}, B.~{Raj}, and L.~{Song}, ``Sphereface:
  Deep hypersphere embedding for face recognition,'' in \emph{2017 IEEE
  Conference on Computer Vision and Pattern Recognition (CVPR)}, 2017, pp.
  6738--6746.

\bibitem{facenet}
F.~Schroff, D.~Kalenichenko, and J.~Philbin, ``Facenet: A unified embedding for
  face recognition and clustering,'' in \emph{Proceedings of the IEEE
  Conference on Computer Vision and Pattern Recognition (CVPR)}, June 2015.

\bibitem{maskedgan1}
C.~Li, S.~Ge, D.~Zhang, and J.~Li, ``Look through masks: Towards masked face
  recognition with de-occlusion distillation,'' in \emph{Proceedings of the
  28th ACM International Conference on Multimedia}, ser. MM '20.\hskip 1em plus
  0.5em minus 0.4em\relax New York, NY, USA: Association for Computing
  Machinery, 2020, p. 3016–3024.

\bibitem{vgg2}
Q.~Cao, L.~Shen, W.~Xie, O.~M. Parkhi, and A.~Zisserman, ``Vggface2: A dataset
  for recognising faces across pose and age,'' 2018.

\bibitem{msceleb}
Y.~Guo, L.~Zhang, Y.~Hu, X.~He, and J.~Gao, ``Ms-celeb-1m: A dataset and
  benchmark for large-scale face recognition,'' in \emph{ECCV}, vol. 9907, 10
  2016, pp. 87--102.

\bibitem{dmonteroTF2}
\BIBentryALTinterwordspacing
D.~Montero, ``face\_recognition\_tf2,'' 2019. [Online]. Available:
  \url{https://github.com/dmonterom/face\_recognition\_TF2}
\BIBentrySTDinterwordspacing

\bibitem{lfw}
G.~Huang, M.~Mattar, T.~Berg, and E.~Learned-Miller, ``Labeled faces in the
  wild: A database forstudying face recognition in unconstrained
  environments,'' \emph{Tech. rep.}, 10 2008.

\bibitem{cfp}
S.~Sengupta, J.~Cheng, C.~Castillo, V.~Patel, R.~Chellappa, and D.~Jacobs,
  ``Frontal to profile face verification in the wild,'' in \emph{IEEE
  Conference on Applications of Computer Vision}, February 2016.

\bibitem{agedb}
S.~Moschoglou, A.~Papaioannou, C.~Sagonas, J.~Deng, I.~Kotsia, and
  S.~Zafeiriou, ``Agedb: the first manually collected, in-the-wild age
  database,'' in \emph{Proceedings of the IEEE Conference on Computer Vision
  and Pattern Recognition Workshop}, vol.~2, 2017, p.~5.

\end{thebibliography}

\end{document}